\documentclass[11pt]{article}
\PassOptionsToPackage{square,numbers}{natbib}
\usepackage{emnlp2016}
\usepackage{url}
\usepackage{latexsym}
\usepackage{amsmath}
\usepackage{amssymb}
\usepackage{color,soul}
\usepackage{booktabs}
\usepackage{graphicx}
\usepackage{lipsum}
\usepackage{nicefrac}       
\usepackage{microtype}      
\usepackage{natbib}
\usepackage{soul}

\usepackage{tikz}
\usetikzlibrary{calc,trees,positioning,arrows,chains,shapes.geometric,%
  decorations.pathreplacing,decorations.pathmorphing,shapes,%
  matrix,shapes.symbols,fit,decorations}
\usepackage{wasysym}
\usepackage{xcolor}
\definecolor{nice-red}{HTML}{E41A1C}
\definecolor{nice-orange}{HTML}{FF7F00}
\definecolor{nice-yellow}{HTML}{FFC020}
\definecolor{nice-green}{HTML}{4DAF4A}
\definecolor{nice-blue}{HTML}{377EB8}
\definecolor{nice-purple}{HTML}{984EA3}
\definecolor{airforceblue}{rgb}{0.36, 0.54, 0.66}
\definecolor{blue}{rgb}{0.2, 0.2, 0.6}
\definecolor{cadmiumgreen}{rgb}{0.0, 0.42, 0.24}

\newcommand{\ie}{{\emph{i.e.}}}
\newcommand{\eg}{{\emph{e.g.}}}

\newcommand{\expremise}[1]{\ul{\tt #1}}
\newcommand{\ex}[1]{{\tt #1}}

\emnlpfinalcopy 
\def\emnlppaperid{***}

\title{Generating Natural Language Inference Chains}

\author{Vladyslav Kolesnyk \and Tim Rockt\"aschel \and Sebastian Riedel\\
      	University College London\\
      	London, UK\\
      	{\small\tt vladyslav.kolesnyk.12@ucl.ac.uk, \{t.rocktaschel,s.riedel\}@cs.ucl.ac.uk}        
 }

\date{}

\begin{document}
\maketitle
\begin{abstract}
The ability to reason with natural language is a fundamental prerequisite for many NLP tasks such as information extraction, machine translation and question answering.
To quantify this ability, systems are commonly tested whether they can recognize textual entailment, \ie{}, whether one sentence can be inferred from another one.
However, in most NLP applications only single source sentences instead of sentence pairs are available. 
Hence, we propose a new task that measures how well a model can \emph{generate} an entailed sentence from a source sentence.
We take entailment-pairs of the Stanford Natural Language Inference corpus and train an LSTM with attention.
On a manually annotated test set we found that $82\%$ of generated sentences are correct, an improvement of $10.3\%$ over an LSTM baseline.
A qualitative analysis shows that this model is not only capable of shortening input sentences, but also inferring new statements via paraphrasing and phrase entailment.
We then apply this model recursively to input-output pairs, thereby generating natural language inference chains that can be used to automatically construct an entailment graph from source sentences. 
Finally, by swapping source and target sentences we can also train a model that given an input sentence invents additional information to generate a new sentence.
\end{abstract}

\section{Introduction}
The ability to determine entailment or contradiction between natural language text is essential for improving the performance in a wide range of natural language processing tasks.
Recognizing Textual Entailment (RTE) is a task primarily designed to determine whether two natural language sentences are independent, contradictory or in an entailment relationship where the second sentence (the hypothesis) can be inferred from the first (the premise). 
Although systems that perform well in RTE could potentially be used to improve question answering, information extraction, text summarization and machine translation~\cite{dagan2006pascal}, only in few of such downstream NLP tasks sentence-pairs are actually available.
Usually, only a single source sentence (\eg{} a question that needs to be answered or a source sentence that we want to translate) is present and models need to come up with their own hypotheses and commonsense knowledge inferences.

The release of the large Stanford Natural Language Inference (SNLI) corpus~\cite{bowman2015large} allowed end-to-end differentiable neural networks to outperform feature-based classifiers on the RTE task~\cite{bowman2016fast,cheng2016long,rocktaschel2015reasoning,vendrov2015order,wang2015learning}. 

In this work, we go a step further and investigate how well recurrent neural networks can produce true hypotheses given a source sentence.
Furthermore, we qualitatively demonstrate that by only training on input-output pairs and recursively generating entailed sentence we can generate natural language inference chains (see Figure \ref{figure:chain_inferred_sentences} for an example).
Note that every inference step is interpretable as it is mapping one natural language sentence to another one.

Our contributions are fourfold:
(i) we propose an entailment generation task based on the SNLI corpus (\S\ref{sec:entailment}),
(ii) we investigate a sequence-to-sequence model and find that $82\%$ of generated sentences are correct (\S\ref{sec:qualitative}),
(iii) we demonstrate the ability to generate natural language inference chains trained solely from entailment pairs (\S\ref{sec:chain}), 
and finally (iv) we can also generate sentences with more specific information by swapping source and target sentences during training (\S\ref{sec:inverse}).

\begin{figure}[t!]
\centering
    \resizebox{\columnwidth}{!}{%
    \begin{tabular}{p{9cm}}
     \expremise{Young blond woman putting her foot into a water fountain} $\rightarrow$ \ex{A person is dipping her foot into water.} $\rightarrow$ \ex{A person is wet.}    
     \end{tabular}
    }
 \caption{Example of a natural language inference chain that is generated from the first sentence.}
 \label{figure:chain_inferred_sentences}
\end{figure}

\section{Method}
In the section, we briefly introduce the entailment generation task and our sequence-to-sequence model.

\subsection{Entailment Generation}
\label{sec:entailment}

To create the entailment generation dataset, we simply filter the Stanford Natural Language Inference corpus for sentence-pairs of the entailment class. 
This results in a training set of $183,416$ sentence pairs, a development set of $3,329$ pairs and a test of $3,368$ pairs. 
Instead of a classification task, we can now use this dataset for a sequence transduction task.

\subsection{Sequence-to-Sequence}

Sequence-to-sequence recurrent neural networks \cite{sutskever2014sequence} have been successfully employed for many sequence transduction tasks in NLP such as machine translation \cite{bahdanau2014neural, cho2014learning},  constituency parsing \cite{vinyals2015grammar}, sentence summarization \cite{rush2015neural} and question answering \cite{hermann2015teaching}.
They consist of two recurrent neural networks (RNNs): an encoder that maps an input sequence of words into a dense vector representation, and a decoder that conditioned on that vector representation generates an output sequence.
Specifically, we use long short-term memory~(LSTM) RNNs \cite{hochreiter1997long} for encoding and decoding.
Furthermore, we experiment with \emph{word-by-word attention} \cite{bahdanau2014neural}, which allows the decoder to search in the encoder outputs to circumvent the LSTM's memory bottleneck. We use greedy decoding at test time.
The success of LSTMs with attention in sequence transduction tasks makes them a natural choice as a baseline for entailment generation, and we leave the investigation of more advanced models to future work.

\subsection{Optimization and Hyperparameters}
We use stochastic gradient descent with a mini-batch size of $64$ and the ADAM optimizer \cite{kingma2014adam} with a first momentum coefficient of $0.9$ and a second momentum coefficient of $0.999$. 
Word embeddings are initialized with pre-trained word2vec vectors \cite{mikolov2013distributed}. Out-of-vocabulary words ($10.5\%$) are randomly initialized by sampling values uniformly from $[-\sqrt3, \sqrt3]$ and optimized during training. 
Furthermore, we clip gradients using a norm of $5.0$. We stop training after $25$ epochs.

\section{Experiments and Results}
We present results for various tasks: (i) given a premise, generate a sentence that can be inferred from the premise, (ii) construct inference chains by recursively generating sentences, and 
(iii) given a sentence, create a premise that would entail this sentence, \ie{}, make a more descriptive sentence by adding specific information.

\subsection{Quantitative Evaluation}
We train an LSTM with and without attention on the training set.
After training, we take the best model in terms of BLEU score~\cite{papineni2002bleu} on the development set and calculate the BLEU score on the test set.
To our surprise, we found that using attention yields only a marginally higher BLEU score (43.1 vs. 42.8).
We suspect that this is due to the fact that generating entailed sentences has a larger space of valid target sequences, which makes the use of BLEU problematic and penalizes correct solutions.
Hence, we manually annotated $100$ random test sentences and decided whether the generated sentence can indeed be inferred from the source sentence. We found that sentences generated by an LSTM with attention are substantially more accurate ($82\%$ accuracy) than those generated from an LSTM baseline ($71.7\%$). To gain more insights into the model's capabilities, we turn to a thorough qualitative analysis of the attention LSTM model in the remainder of this paper.

\subsection{Example Generations}
\label{sec:qualitative}
\begin{figure}[t!]
    \centering
    \resizebox{\columnwidth}{!}{%
    \begin{tabular}{p{9cm}}
        \begin{enumerate}
          \item \label{synt_simpl} \expremise{He is driving a beautiful red car.} $\rightarrow$  \ex{A man is driving a car.}          
          \item \label{numerals} \expremise{Two men and a woman finishing a meal and drinks.}  $\rightarrow$ \ex{Three people are eating together.}
          \item \label{knowledge_infer} \expremise{A father holds a new born baby.}  $\rightarrow$ \ex{A man holds his baby.} 
        \end{enumerate} 
     \end{tabular}
    }
 \caption{Valid sentences generated by the model.}
 \label{figure:inferred_sentences_examples}
\end{figure}
Figure \ref{figure:inferred_sentences_examples} shows examples of generated sentences from the development set.
Syntactic simplification of the input sentence seems to be the most common approach. 
The model removes certain parts of the premise such as adjectives, resulting in a more abstract sentence (see Figure \ref{figure:inferred_sentences_examples}.\ref{synt_simpl}).

Figure \ref{figure:inferred_sentences_examples}.\ref{numerals} demonstrates that the system can 
recognize the number of subjects in the sentence and includes this information in the generated sentence. 
However, we did not observe such 'counting' behavior for more than four subjects, indicating that the system memorized frequency patterns from the training set. 

Furthermore, we found predictions that hint to common-sense assumptions: if a sentence talks about a father holding a newborn baby, it is most likely that the newborn baby is his own child
(Example \ref{figure:inferred_sentences_examples}.\ref{knowledge_infer}).

\paragraph{Limitations}
\begin{figure}[t]
    \centering
    \resizebox{\columnwidth}{!}{%
    \begin{tabular}{p{9cm}}
        \begin{enumerate}
          \item \label{synt_simpl} \expremise{A female gymnast in black and red being coached on bar skills.} $\rightarrow$  \ex{A female is in a bar.}      
          \item \label{tense_change} \expremise{A dog drops a yellow disc on a beach.} $\rightarrow$ \ex{The dog drops a red disc.}
        \end{enumerate} 
     \end{tabular}
    }
 \caption{Invalid sentences generated by the model.}
 \label{figure:invalid_examples}
\end{figure}
Two reappearing limitations of the proposed model are related to dealing with words that have a very different meaning but similar word2vec embeddings (\eg{} colors), as well as ambiguous words.
For instance, 'bar' in Figure \ref{figure:invalid_examples}.\ref{synt_simpl} refers to pole vault and not a place in which you can have a drink.
Substituting one color by another one (Figure \ref{figure:invalid_examples}.\ref{tense_change}) is a common mistake.

\paragraph{Out-of-Corpus Examples}
\begin{figure}[t!]
    \centering
    \resizebox{\columnwidth}{!}{%
    \begin{tabular}{p{9cm}}
        \begin{enumerate}
          \item \label{go_example} \expremise{First time a computer Go program has beaten a 9-dan professional without handicaps.}  $\rightarrow$ \ex{A person is using a computer.} 
          \item \label{alter_1} \expremise{Competitors must be dressed in suitable tennis attire that is almost entirely white.}  $\rightarrow$ \ex{The tennis player is wearing white.} 
          \item \label{alter_4} \expremise{The athletes run by the harbour during the Haspa Hamburg marathon.}  $\rightarrow$ \ex{The runners are outside.} 
          \item \label{alter_6} \expremise{Space travellers gaze up and strain to find the blue dot in their sky.}  $\rightarrow$ \ex{There are people looking at the sky.} 
        \end{enumerate} 
     \end{tabular}
    }
 \caption{Generated sentences from external sources: (1) Wikipedia, (2) wimbledon.com, (3) a news article, (4) "The Frontier is Everywhere" by Carl Sagan.}
 \label{figure:inferred_outside_dev_examples}
\end{figure}

The SNLI corpus might not reflect the variety of sentences that can be encountered in downstream NLP tasks. 
In Figure \ref{figure:inferred_outside_dev_examples} we present generated sentences for randomly selected examples of out-of-domain textual resources.
They demonstrate that the model generalizes well to out-of-domain sentences, making it a potentially very useful component for improving systems for question answering, information extraction, sentence summarization etc.

\subsection{Inference Chain Generation}
\label{sec:chain}
\begin{figure}[t!]
    \centering
    \resizebox{\columnwidth}{!}{%
    \begin{tabular}{p{9cm}}
        \begin{enumerate}       
           \item \label{chain_2sentence_3} 
           \expremise{People are celebrating a victory on the square.} 
           $\rightarrow$ \ex{People are celebrating.} 
           $\rightarrow$ \ex{People are happy.} 
           $\rightarrow$
           \ex{\hl{The people are smiling.}}
           $\leftarrow$ \ex{People are smiling.}
           $\leftarrow$ \ex{Some people smiling.}
           $\leftarrow$ \ex{couple smiling}
           $\leftarrow$ \ex{A bride and groom smiles.}
           $\leftarrow$ \expremise{A wedding party looks happy on the picture.}           
           \item \label{chain_2sentence_4} 
           \expremise{The athletes run at the start of the Haspa Hamburg marathon.} $\rightarrow$ \ex{The runners are at the race.} $\rightarrow$ \ex{There are runners at the race.} $\rightarrow$
           \ex{There are people running.} $\rightarrow$
           \ex{\hl{People are running.}}
           $\leftarrow$ \ex{People are running outside.}
           $\leftarrow$ \expremise{People are running away from the bear.}
        \end{enumerate} 
     \end{tabular}
    }
 \caption{Examples of inference chains where two premises (underlined) converge to the same sentence (highlighted).}
 \label{figure:chain_example}
\end{figure}

Next, we test how well the model can generate inference chains by repeatedly passing generated output sentences as inputs to the model.
We stop once a sentence has already been generated in the chain.
Figure \ref{figure:chain_example} shows that this works well despite that the model was only trained on sentence-pairs.

Furthermore, by generating inference chains for all sentences in the development set we construct an entailment graph. 
In that graph we found that sentences with shared semantics are
eventually mapped to the same sentence that captures the shared meaning.

\begin{figure}[t!]
  \centering
  \resizebox{0.9\columnwidth}{!}{%
  \includegraphics{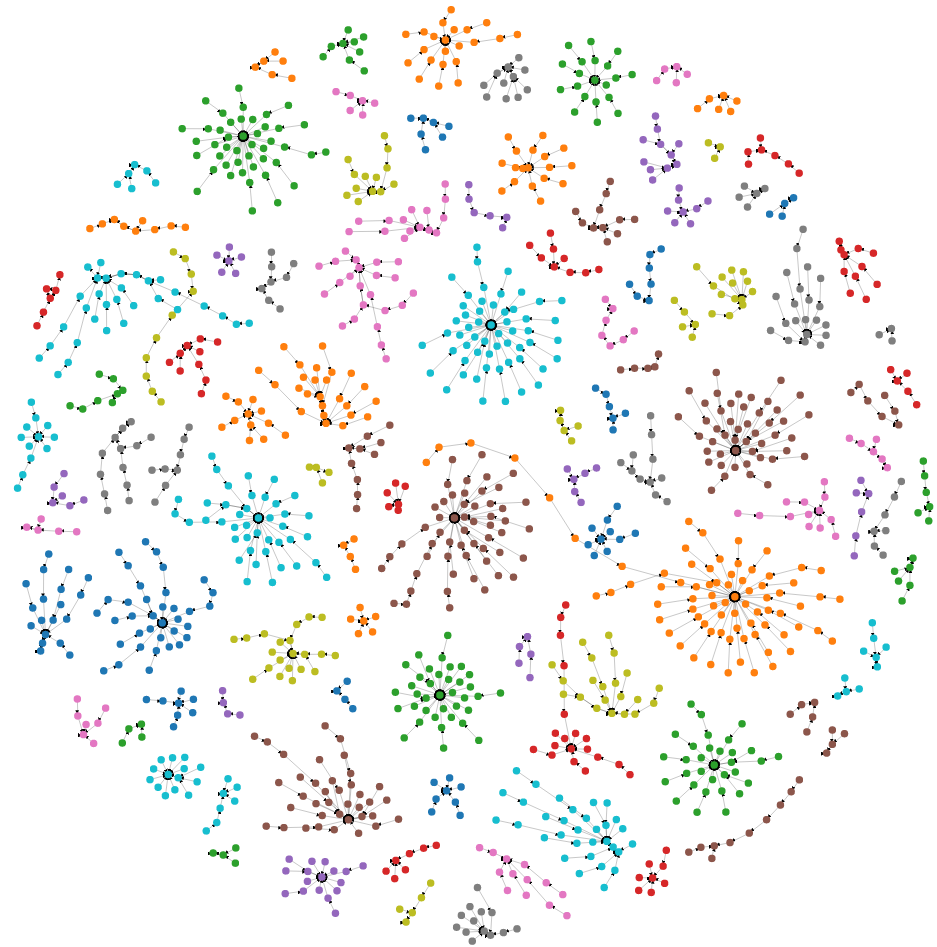}
  }
  \caption{Topology of the generated entailment graph.}
  \label{figure:big_vocab_network}
\end{figure}

A visualization of the topology of the entailment graph is shown in Figure \ref{figure:big_vocab_network}. 
Note that there are several long inference chains, as well as large clusters of sentences (nodes) that are mapped (links) to the same shared meaning.

\subsection{Inverse Inference}
\label{sec:inverse}
\begin{figure}[t!]
    \centering
    \resizebox{\columnwidth}{!}{%
    \begin{tabular}{p{9cm}}
        \begin{enumerate}         
          \item \label{inv_1} \expremise{Two girls sit by a lake.} $\rightarrow$ \ex{Two girls are sitting on a ledge in front of a lake.}
          
          \item \label{inv_3} 
          \expremise{Two girls are playing chess} $\rightarrow$ 
          \ex{Two girls playing chess in a restaurant.}
          
          \item \label{inv_chain_2} 
          \expremise{A woman is outside.} $\rightarrow$ 
          \ex{A woman is walking down a street in a city.} $\rightarrow$ 
          \ex{A woman in a black skirt walks down a city street.}                
        \end{enumerate} 
     \end{tabular}
    }
 \caption{Generated sentences for the inverse task.}
 \label{figure:inverse_chains_examples}
\end{figure}

By swapping the source and target sequences for training, we can train a model that given a sentence invents additional information to generate a new sentence (Figure \ref{figure:inverse_chains_examples}).
We believe this might prove useful to increase the language variety and complexity of AI unit tests such as the Facebook bAbI task \cite{weston2015towards}, but we leave this for future work.

\section{Conclusion and Future Work}
We investigated the ability of sequence-to-sequence models to generate entailed sentences from a source sentence.
To this end, we trained an attentive LSTM on entailment-pairs of the SNLI corpus. 
We found that this works well and generalizes beyond in-domain sentences.
Hence, it could become a useful component for improving the performance of other NLP systems.

We were able to generate natural language inference chains by recursively generating sentences from previously inferred ones.
This allowed us to construct an entailment graph for sentences of the SNLI development corpus.
In this graph, the shared meaning of two related sentences is represented by the first natural language sentence that connects both sentences.
Every inference step is interpretable as it maps a natural language sentence to another one.

Towards high-quality data augmentation, we experimented with reversing the generation task. 
We found that this enabled the model to learn to invent specific information.

For future work, we want to integrate the presented model into larger architectures to improve the performance of downstream NLP tasks such as information extraction and question answering.
Furthermore, we plan to use the model for data augmentation to train expressive neural networks on tasks where only little annotated data is available.
Another interesting research direction is to investigate methods for increasing the diversity of the generated sentences.

\clearpage

\section*{Acknowledgments}
We thank Guillaume Bouchard for suggesting the reversed generation task, and Dirk Weissenborn, Isabelle Augenstein and Matko Bosnjak for comments on drafts of this paper. This work was supported by Microsoft Research through its PhD Scholarship Programme, an Allen Distinguished Investigator Award, and a Marie Curie Career Integration Award. 

\bibliography{nlci}
\bibliographystyle{abbrvnat}

\appendix

\end{document}